\documentclass[runningheads]{llncs}

\usepackage{eccv}
\usepackage{multirow}
\usepackage{wrapfig}
\usepackage{enumitem}
\usepackage[table]{xcolor} 
\usepackage{array}
\usepackage{booktabs}
\usepackage{dsfont}

\usepackage{eccvabbrv}

\usepackage{graphicx}
\usepackage{booktabs}

\usepackage[accsupp]{axessibility}  

\usepackage{hyperref}

\usepackage{orcidlink}

\begin{document}

\title{DoReMi: Bridging 3D Domains via Topology-Aware Domain-Representation Mixture of Experts} 

\titlerunning{DoReMi}



\author{Mingwei Xing$^*$ \and
Xinliang Wang$^*$ \and
Yifeng Shi$^\dagger$}
\authorrunning{M.~Xing et al.}

\institute{Ke Holdings Inc.\\
\email{\{xingmingwei001, wangxinliang008, shiyifeng003\}@ke.com}}

{\let\thefootnote\relax\footnotetext{$^*$Equal contribution \quad $^\dagger$ Corresponding author}}
\maketitle

\begin{abstract}
Constructing a unified 3D scene understanding model has long been hindered by the significant topological discrepancies across different sensor modalities. While applying the Mixture-of-Experts (MoE) architecture is an effective approach to achieving universal understanding, we observe that existing 3D MoE networks often suffer from semantics-driven routing bias. This makes it challenging to address cross-domain data characterized by ``semantic consistency yet topological heterogeneity.'' To overcome this challenge, we propose \textbf{DoReMi} (Topology-Aware Domain–Representation Mixture of Experts). Specifically, we introduce a self-supervised pre-training branch based on multi attributes, such as topological and texture variations, to anchor cross-domain structural priors. Building upon this, we design a domain-aware expert branch comprising two core mechanisms: Domain Spatial-Guided Routing (DSR), which achieves an acute perception of local topological variations by extracting spatial contexts, and Entropy-controlled Dynamic Allocation (EDA), which dynamically adjusts the number of activated experts by quantifying routing uncertainty to ensure training stability. Through the synergy of these dual branches, DoReMi achieves a deep integration of universal feature extraction and highly adaptive expert allocation. Extensive experiments across various tasks, encompassing both indoor and outdoor scenes, validate the superiority of DoReMi. It achieves 80.1\% mIoU on the ScanNet validation set and 77.2\% mIoU on S3DIS, comprehensively outperforming existing state-of-the-art methods. The code will be released soon.
\end{abstract}

\section{Introduction}
\label{sec:intro}
Driven by the demand for universal 3D perception, leveraging massive multi-source data for unified scene understanding has become a prominent trend~\cite{ppt, wang2024one, chen2025point,sonata}. However, unlike 2D images which are primarily influenced by illumination and texture, the 3D domain faces more severe challenges from sensor heterogeneity. Different sensors, such as LiDAR ray scanning and RGB-D or Mesh surface reconstruction, cause the same semantic object, such as a wall, to exhibit distinct physical topological structures ranging from sparse line patterns to dense surfaces. This semantically consistent yet topologically heterogeneous nature makes it difficult for models to reconcile conflicting geometric patterns during joint training, which subsequently leads to feature space degradation and negative transfer.
\begin{figure*}[t]
    \centering
    \includegraphics[width=0.9\linewidth]{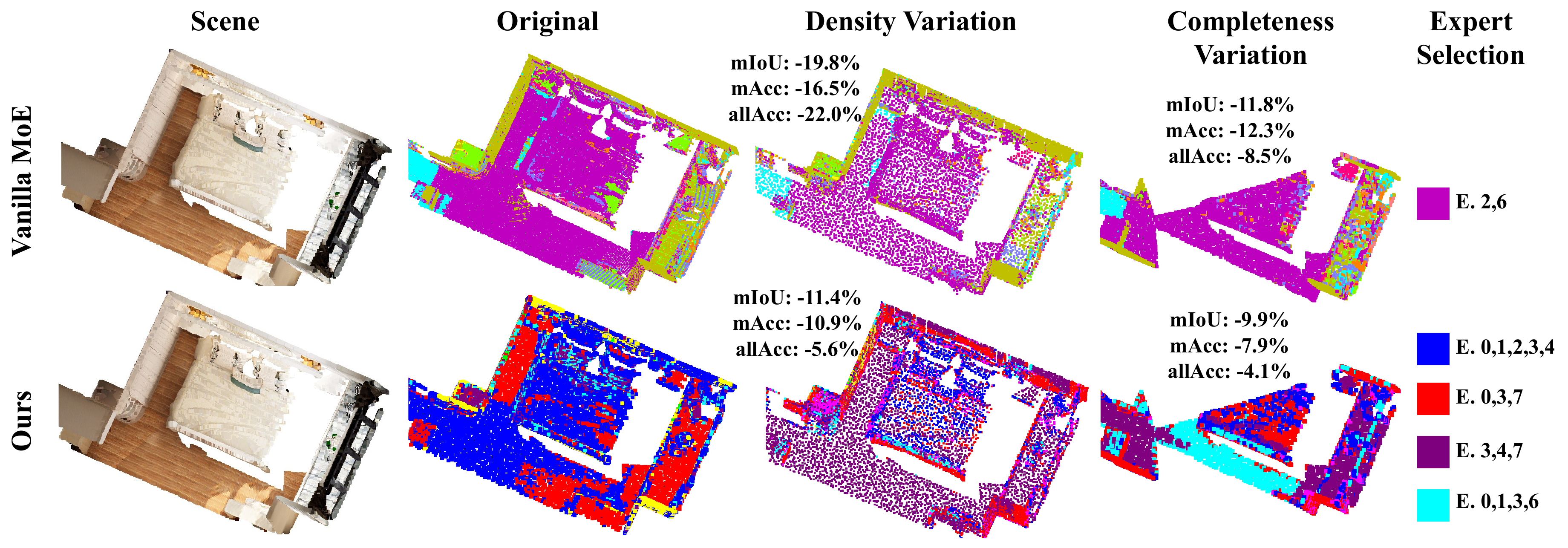}
    \caption{\textbf{Semantic-dominant vs. Topology-aware routing.} Existing 3D MoE methods (e.g., Point-MoE~\cite{chen2025point}), referred to as Vanilla MoE, rely on rigid, semantic-driven routing. When an identical semantic object (e.g., ``bed'' or ``floor'') undergoes topological changes such as density or completeness variations, Vanilla MoE rigidly assigns the exact same experts, suffering severe performance drops (e.g., a 19.8\% mIoU drop under density variation). In contrast, DoReMi dynamically allocates specialized expert subsets to handle these fine-grained geometric changes, significantly mitigating performance degradation and ensuring robust 3D understanding.}
    \label{fig:teaser}
\end{figure*}
\begin{wrapfigure}{r}{0.4\textwidth}
    \centering
    \includegraphics[width=1.0\linewidth]{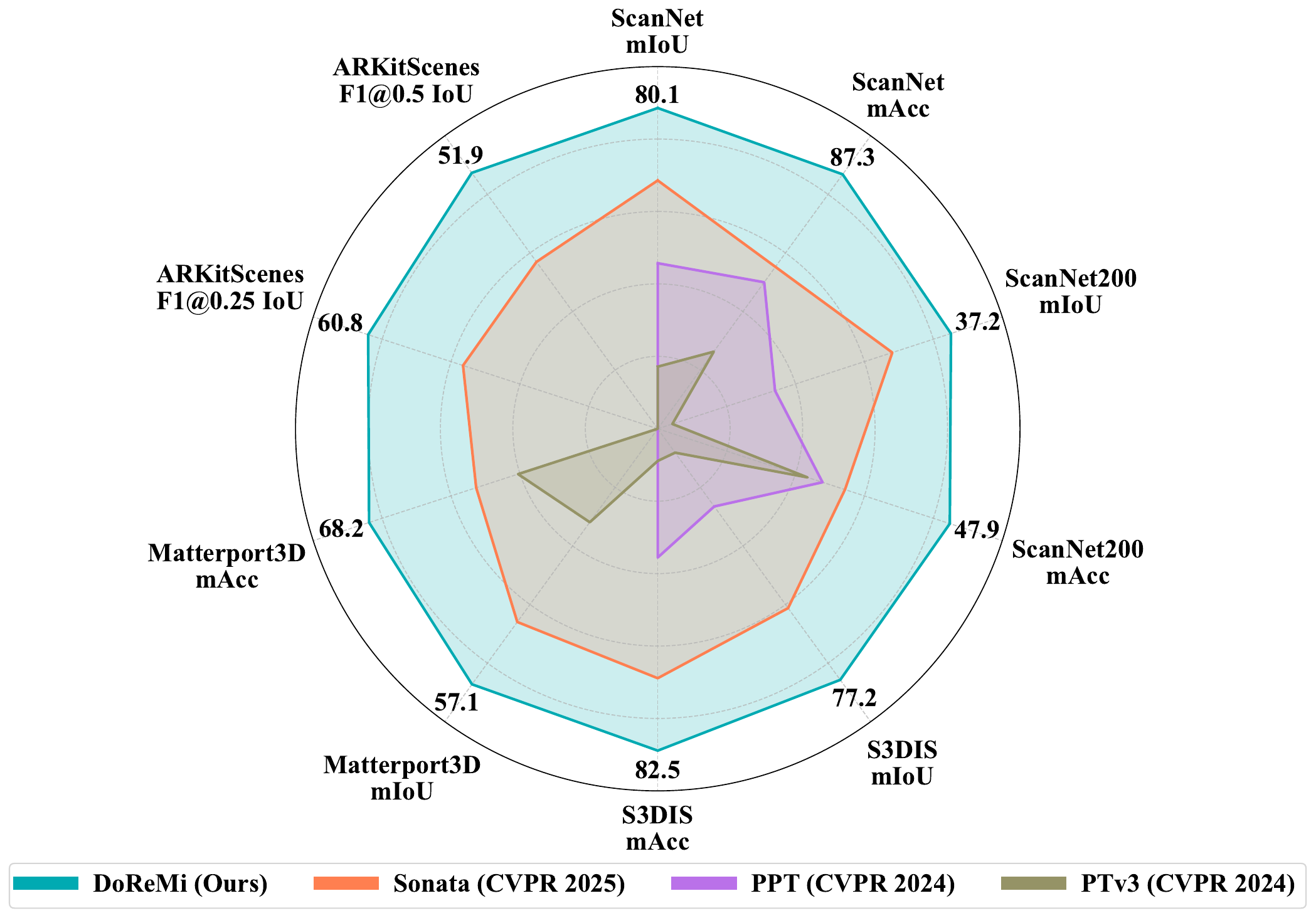}
    \caption{\textbf{Performance of DoReMi}. DoReMi advances multiple SOTA benchmarks on 3D scene understanding datasets.}
    \label{fig:radar}
\end{wrapfigure}

To address this challenge, existing joint training paradigms primarily explore two categories of methods: unified representation learning and modularized adaptation. Unified representation learning aims to construct a single feature space to align multi-source data~\cite{pointcontrast,pointmae,sonata}. However, such forced alignment often results in geometric smoothing, where the model suppresses sensor-specific geometric details to fit the global cross-domain distribution, thereby compromising precision. In contrast, modularized adaptation paradigms attempt to maintain flexibility by introducing specialized modules into a shared backbone, yet they are limited by insufficient modeling of 3D heterogeneity. Static strategies such as~\cite{ppt,wang2024one} can only learn globally fixed parameters and do not respond dynamically to drastic variations in density within 3D point clouds. Although dynamic strategies such as 3D Mixture of Experts (MoE)~\cite{chen2025point,uni3dmoe,xu2025limoe} introduce routing mechanisms, they typically follow the semantic-dominant routing common in natural language processing (illustrated in Figure~\ref{fig:teaser} and are further discussed in Section~\ref{sec:experiments})). This mechanism assigns experts based solely on object semantics and overlooks the fact that 3D domain gaps originate from geometric topology rather than semantics. When the same semantic object presents different topological structures under different sensors, routing based purely on semantics cannot distinguish or adapt to these physical differences, leading to accelerated performance degradation.

Thus, we propose DoReMi, a framework consisting of two distinct branches. Specifically, the representation branch (Re) leverages self-supervised pre-training based on multiple attributes, such as topological and textural variations, to provide the model with rich and robust feature representations. Complementing this, the domain-aware branch (Do) is designed as a topology-aware Mixture-of-Experts (MoE) to resolve topological inconsistencies arising from sensor heterogeneity. The core of this approach lies in the synergy between Domain-Spatial-Guided Routing (DSR) and Entropy-controlled Dynamic Allocation (EDA). First, DSR enables the model to perceive topological variations by extracting spatial context. Second, EDA is introduced to maintain a stable and balanced distribution of experts, thereby ensuring training robustness. By utilizing the Do branch, the model dynamically activates the most suitable combination of experts for specific data distributions. Through the collaboration of these two branches, DoReMi effectively handles the inherent topological fluctuations in multi-source 3D data, enhancing overall 3D understanding capabilities. As illustrated in Figure \ref{fig:radar}, DoReMi achieves superior performance across multiple 3D understanding tasks compared to existing state-of-the-art methods. Our main contributions are summarized as follows:
\begin{itemize}[label=$\bullet$]
\item \textbf{Identification and resolution of semantic bias.} We uncover a critical limitation: existing 3D MoEs often rely on semantic-driven routing, which consistently overlooks sensor-induced topological variations. To resolve this, DoReMi pairs robust representations with topology-aware routing, balancing cross-domain invariance with sensitivity to fine-grained geometric details.
\item \textbf{Synergistic design of dual-branch framework.} DoReMi decouples representation learning into two mutually reinforcing branches. The frozen Re branch employs self-supervised augmentations simulating physical variations to filter domain noise and establish a robust cross-domain structural anchor. Grounded in this prior, the Do branch leverages DSR and EDA for highly adaptive, topology-aware expert allocation.

\item \textbf{Comprehensive evaluation and SOTA performance.} Extensive experiments on multiple indoor and outdoor 3D understanding benchmarks demonstrate that DoReMi consistently outperforms existing approaches. Specifically, it achieves 80.1\% mIoU on ScanNet Val and 77.2\% mIoU on S3DIS, establishing a new state of the art in generalizable 3D understanding.
\end{itemize}

\section{Related Work}
\subsection{3D Understanding}
3D scene understanding is fundamental to computer vision, covering tasks such as semantic segmentation~\cite{qi2017pointnet, qi2017pointnet++, thomas2019kpconv, pointcontrast, qian2022pointnext}, object detection~\cite{zhou2018voxelnet, lang2019pointpillars, shi2020pv}, and instance segmentation~\cite{jiang2020pointgroup, he2022pointinst3d}. While early voxel-based methods faced scalability limits, modern point-based~\cite{qi2017pointnet++, thomas2019kpconv} and transformer-based~\cite{ptv2, ptv3} architectures have significantly improved performance. However, most models remain domain-specific and lack robust cross-domain generalization. 
In this context, our proposed DoReMi framework can serve as a universal 3D feature extractor, capable of providing both domain-adaptive and cross-domain generalizable representations, thereby enhancing the performance of various downstream 3D understanding tasks.

\subsection{Unified and Adaptive 3D Representation Learning}
Building a universal 3D scene understanding model primarily involves a trade-off between generalization and domain specificity. Current joint training paradigms can be categorized into two approaches. Unified representation learning constructs a shared feature space through large-scale self-supervised pre-training~\cite{pointcontrast, pointmae,sonata, zhou2023uni3d,zhang2025concerto,kolodiazhnyi2025unidet3d,soum2023mdt3d}. Although this enhances generalization, the forced alignment of distributions frequently triggers geometric smoothing, which leads to the loss of discriminative details specific to certain sensors. Conversely, modularized adaptation paradigms~\cite{beyondsparse,ppt,wang2024one} attempt to maintain flexibility by introducing specialized parameters. However, their static strategies fail to respond to the drastic local topological fluctuations inherent in 3D data. Distinct from these methods, DoReMi abandons forced alignment in favor of DSR. By dynamically perceiving local geometric topology, DoReMi enables adaptive fine-grained modeling across heterogeneous point clouds while maintaining strong global priors.
\begin{figure*}[t]
    \centering
    \includegraphics[width=0.98\linewidth]{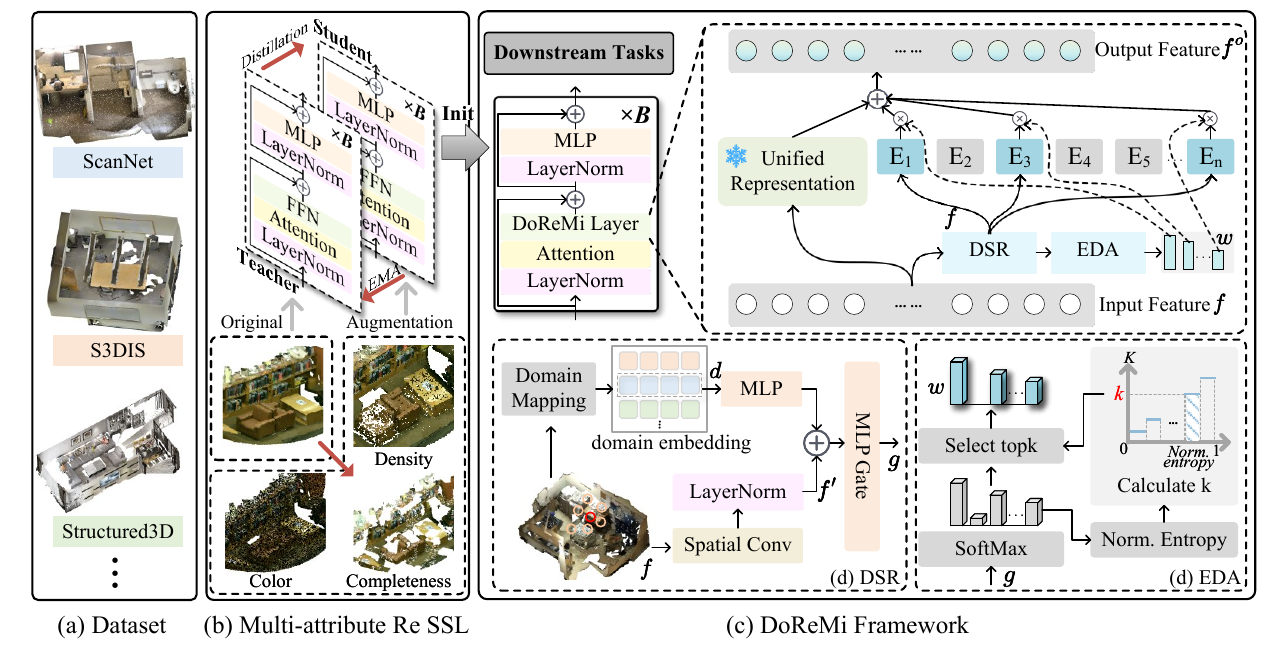}
    \caption{\textbf{The architecture of DoReMi.} It combines two branches: a domain-aware branch (Do) and a unified-representation branch (Re). DoReMi starts with a pretrained model generated through self-supervised learning on multi-attribute data. Re freezes this pretrained knowledge to extract cross-domain geometric and structural patterns. Meanwhile, Do uses Domain-Spatial-Guided Routing (DSR) and Entropy-Controlled Dynamic Allocation (EDA) to capture domain-specific features. Together, these branches enhance both domain adaptability and generalization.}
    \label{fig:framework}
\end{figure*}
\subsection{Mixture-of-Experts}
Mixture of Experts (MoE) has been extensively applied to address multi-domain joint training and domain generalization. In 2D vision, \cite{li2022sparse,dai2021generalizable,xu2024cbdmoe,zhong2022meta} integrate experts into the Transformer architecture to achieve effective adaptation for large-scale image distributions. Recently, research has begun to explore the application of MoE in the 3D domain, including Point-MoE \cite{chen2025point},Uni3D-MoE \cite{uni3dmoe}, LiMoE \cite{xu2025limoe}. However, most of these methods adopt routing logic based on task labels or semantic features, which fails to perceive the physical topological conflicts inherent in 3D data. These approaches rely primarily on semantic features for routing assignment and struggle to respond to the underlying differences in physical topology. In contrast, DoReMi introduces DSR, enabling dynamic perception and adaptive scheduling across heterogeneous sensor distributions.
\section{Method}

\subsection{Overview}
We propose DoReMi (Domain–Representation Mixture), a Mixture-of-Experts (MoE) architecture for generalizable 3D understanding. It jointly models domain-aware expert features and unified representation features, enabling adaptive modeling of diverse domain distributions while maintaining cross-domain consistency and generalization capability. The overall architecture is illustrated in Figure \ref{fig:framework}. Specifically, we first employ a teacher-student network framework \cite{mean_teacher} to conduct multi-attribute self-supervised learning across multiple datasets, thereby learning geometric domain-structure priors that generalize across domains. Subsequently, the DoReMi architecture undergoes multi-domain supervised joint training, with its weights initialized from the student network. 
FFN weights are duplicated for both the Re and Do branches, while the Re branch remains frozen. Do branch consists of two components: Domain-Spatial-Guided Routing (DSR) and Entropy-Controlled Dynamic Allocation (EDA). DSR leverages local geometric cues to enable the model to perceive topological variations, while EDA maintains the stability of the system by balancing the distribution of experts. By integrating these two mechanisms, the model dynamically activates the most appropriate combination of experts tailored to specific data distributions.
\subsection{Unified Representation Branch}
We introduce a frozen Unified Representation branch (Re), constructed via large-scale multi-attribute self-supervised learning to provide rich and robust feature representations. Through systematic comparison of point cloud data across different domains, we observe significant distributional discrepancies, primarily manifested in color distribution patterns, variations in point cloud spatial density, and object incompleteness caused by occlusions and camera viewpoint limitations. These differences lead to a notable degradation in model generalization performance on cross-domain data. 

After the above analysis, we design three self-supervised alignment tasks tailored to color, density, and completeness, respectively. Specifically, we partition the point cloud into multiple patches. Within each patch, points are randomly assigned black coloration at varying ratios and probabilities, while certain points are also randomly discarded. Additionally, we apply masking operations to entire patches to simulate variations in point cloud completeness. Inspired by DINOv2 \cite{dinov2} and Sonata \cite{sonata}, we adopt a feature distillation framework based on a teacher-student network architecture, where the teacher's weights are updated via an exponential moving average (EMA) of the student's weights. The teacher receives raw data, while the student receives augmented data. The student is trained to synchronize with teacher through a cluster-based loss \cite{caron2020unsupervised,sablayrolles2018spreading}, ensuring uniform feature consistency across diverse augmentations of the same point cloud. This approach effectively suppresses domain-specific noise, thereby providing stable and robust foundational features for domain-specific experts. Furthermore, we initialize the DoReMi architecture with the weights of the student network, duplicating its FFN for all experts in both Re and Do branches, while keeping Re branch frozen. Given point-level tokens $f \in \mathbb{R}^{N \times D}$, where $N$ denotes the number of tokens and $D$ denotes the feature dimension. The output of Re branch is: $f^{\text{Re}} = E_{Re}(f),$
where $E_{Re}$ denotes the expert in the Re branch.
\subsection{Domain-Spatial-Guided Routing}
To achieve topology-aware expert selection, we design DSR. 
Given $f$ and its corresponding domain embedding $d$, we first reshape ${f}$ into a 3D sparse tensor to capture its local topological structure and positional correlations in the spatial dimensions, and then apply a 3D spatial convolution operation for feature extraction. Through a set of learnable convolutional kernels and normalization, the model effectively extracts feature representations ${f}'$ with spatial locality awareness. 
Subsequently, we map the current scene to the corresponding domain embedding $d$ based on its dataset affiliation, and a lightweight MLP then transforms $d$ into a continuous vector $\mathbf{e}_d \in \mathbb{R}^D$ for channel alignment, which encodes domain-specific semantic priors. We add ${e}_d$ to the spatially convolved feature ${f}'$ via broadcasting to generate the domain-aware routing input: ${z} = f' + e_d,$
where $z \in \mathbb{R}^{N\times D}$.This fusion operation ensures that routing decisions depend not only on the input content but are also explicitly guided by the semantic information of the target domain, thereby enhancing the domain adaptability of expert assignment.

The resulting ${z}$ is then fed into a gating network $\mathcal{G}$, composed of a MLP with nonlinear activation functions. $\mathcal{G}$ outputs routing logits: ${g} = \mathcal{G}(z) \in \mathbb{R}^{ N \times K},$
where $K$ is the number of experts.
\subsection{Entropy-Controlled Dynamic Allocation}
To achieve robust and adaptive expert allocation, we propose EDA.
First, we apply softmax function to the gating outputs in last dimension to obtain a probability distribution $p = \text{SoftMax}(g) \in \mathbb{R}^{ N \times K}$, and calculate the Shannon entropy for each token:
\begin{equation}
    H = -\sum_{j=1}^K p[:,j] \odot \log p[:,j],
\end{equation}
where $\odot$ denotes element-wise multiplication and $H\in \mathbb{R}^{ N }$ reflects the model's decision-making uncertainty for that token.
Next, we linearly map the entropy values to the number of experts $k$ to dynamically determine the number of activated experts:

\begin{equation}
k = \left\lceil k_{\min} + \frac{H}{H_{\max}} \cdot (k_{\max} - k_{\min}) \right\rceil,
\end{equation}
where $k\in \mathbb{R}^{ N}$ represents the number of selected experts, $H_{\max} = \log K$ represents the theoretical maximum entropy, with $k_{\min}=1$ and $k_{\max}=K$, and $ \left\lceil \cdot \right\rceil$ denotes the ceiling function. 
Tokens with higher entropy (uncertainty) activate more experts to enhance representation capacity, whereas low-entropy tokens activate fewer experts to improve computational efficiency. 
We sort the experts in descending order of their probabilities $ p $, and activate the top-$ k $ experts with the highest probabilities. Each expert's weight assignment is given by:
\begin{equation}
w[i,j] =
\begin{cases}
p[i,j], & j \in E_i^{\text{act}} \\
0, & otherwise,
\end{cases}
\end{equation}
where $ E_i^{\text{act}} $ denotes the activated expert indices for i-th token and $w \in \mathbb{R}^{N\times K}$.

\begin{table*}[t]
\small
\centering
\caption{\textbf{Indoor semantic segmentation.}}
\begin{tabular}{l *{10}{c}}
\toprule
\multirow{2}{*}{Method}&\multirow{2}{*}{Source}& \multicolumn{3}{c}{ScanNet Val} & \multicolumn{3}{c}{ScanNet200 Val} & \multicolumn{3}{c}{S3DIS Area 5} \\
\cmidrule(lr){3-5} \cmidrule(lr){6-8} \cmidrule(lr){9-11}
 && mIoU & mAcc & allAcc & mIoU & mAcc & allAcc & mIoU & mAcc & allAcc \\
\hline
PTv3 \cite{ptv3} & CVPR 2024& 77.6 & 85.0 & 92.0 & 35.3 & 46.0 & 83.4 & 73.4 & 78.9 & 91.7 \\

CDSegNet \cite{cdsegnet} & CVPR 2025  & 77.9 & 85.2 & 92.2 & 36.3 & 45.9 & 83.9 & - & - & - \\
PPT \cite{ppt} & CVPR 2024& 78.6 & 85.9 & 92.3 & 36.0 & 46.2 & 83.8 & 74.3 & 80.1 & 92.0 \\
Sonata \cite{sonata} & CVPR 2025& 79.4 & 86.1 & 92.5 & 36.8 & 46.5 & \textbf{84.4} & 76.0 & 81.6 & 93.0 \\
\rowcolor[HTML]{caeef0} DoReMi (Ours) & - & \textbf{80.1} & \textbf{87.3} & \textbf{93.1} & \textbf{37.2} & \textbf{47.9} & \textbf{84.4} & \textbf{77.2} & \textbf{82.5} & \textbf{93.1} \\
\bottomrule
\end{tabular}
\label{tab:results}
\end{table*}

To prevent expert imbalance, we apply a load balancing loss\cite{switch_transformer}:
\begin{equation}
\begin{split}
    \mathcal{L}_{\text{balance}} &=  {K} \cdot \sum_{j=1}^K c_j \cdot  r_j ,\\
    c_j = \frac{1}{N} \sum_{i=1}^{N} 	
\mathds{1} \{j &\in E_i^{act}\}, 
    r_j = \frac{1}{N} \sum_{i=1}^{N} p[i,j],
\end{split}
\end{equation}
where $c_j$ is the proportion of tokens routed to expert $j$, and $r_j$ is the average probability assigned by DSR via softmax function. This loss promotes uniform routing, fostering collaborative learning across all experts.

Finally, the Do branch output is computed as a weighted sum of the top-$k$ experts selected by routing probabilities. 
\begin{equation}     
    f^{\text{Do}} = \sum_{j=1}^{K} w[:, j] \odot E_j(f).
\end{equation}
The domain-aware branch accurately captures spatial contexts for targeted expert selection, ensuring stable, robust, and adaptive allocation across diverse domains. 
The output feature is computed as: $f^{o} = f^{\text{Do}} + f^{\text{Re}}.$

\subsection{Training Recipe}
First, we obtain a pretrained model through multi-attribute self-supervised learning and use student model's weights to initialize the parameters of the DoReMi network. 
Subsequently, we adopt multi-dataset joint training as in PPT~\cite{ppt}, unifying category representations through a CLIP-head and InfoNCE loss~\cite{infonce}. The training loss is:
\begin{equation}
    \mathcal{L}_{\text{joint}} = \mathcal{L}_{\text{InfoNCE}} + \lambda \mathcal{L}_{\text{balance}}.
\end{equation}

After completing the joint training, the resulting model can be directly deployed for the primary tasks within the joint training framework and further fine-tuned to address diverse downstream tasks on novel datasets. 

Taking semantic segmentation and multimodal detection as examples. The segmentation loss is defined as the standard cross-entropy loss:
\begin{equation}
\mathcal{L}_{\text{seg}} = -\frac{1}{M} \sum_{i=1}^{M} \sum_{c=1}^{C} y_{i,c} \log(q_{i,c}),
\end{equation}
where $M$ denotes the number of samples, $C$ is the number of classes, $y_{i,c}$ represents the ground-truth label, and $q_{i,c}$ is the predicted probability for class $c$ of sample $i$.

For the multi-modal detection task, we follow SpatialLM \cite{SpatialLM} and leverage the autoregressive property of the Qwen2.5 language model \cite{qwen2} to treat coordinate prediction as a sequence generation task. We adopt the same standard cross-entropy loss as used in Qwen \cite{qwen2}:
\begin{equation}
\mathcal{L}_{\text{det}} = -\frac{1}{T} \sum_{t=1}^{T} \log P(w_t | w_{<t}, \theta),
\end{equation}
where $T$ is the length of the coordinate sequence, $w_t$ denotes the $t$-th coordinate element in the sequence, $w_{<t}$ represents the history of previously generated coordinates, and $\theta$ denotes the model parameters.



\begin{figure*}[t]
    \centering
    \includegraphics[width=0.9\linewidth]{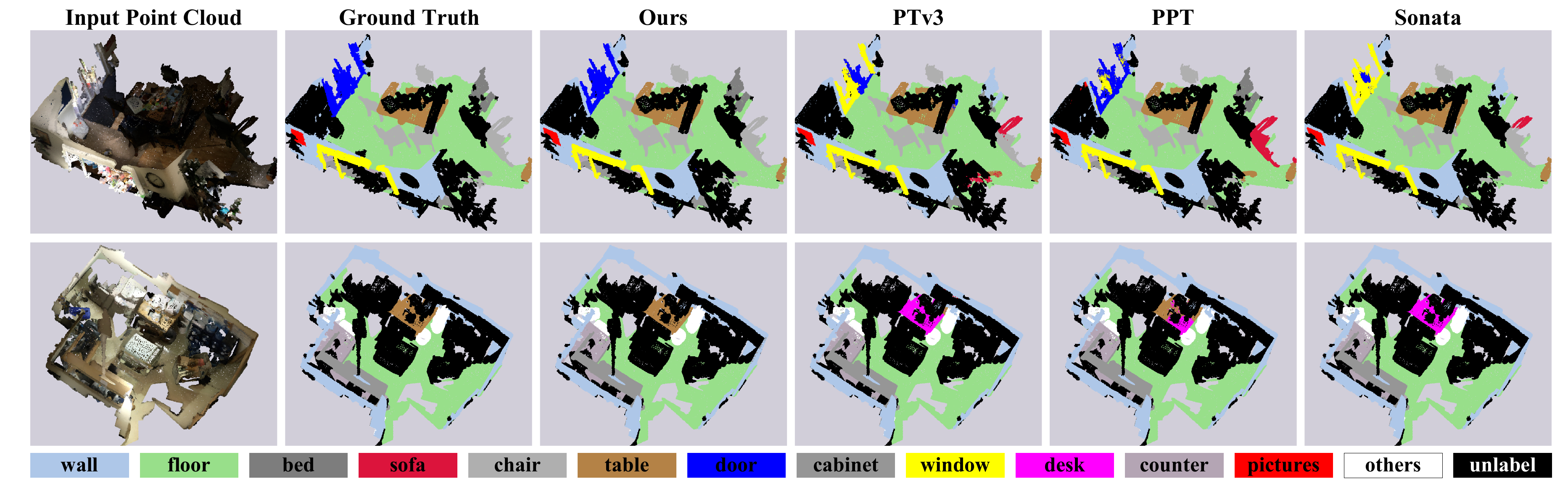}
    \caption{\textbf{Qualitative analysis.} Visualization of different methods on ScanNet.}
    \label{fig:visualizaiton}
\end{figure*}
The overall fine-tune training objective is:

\begin{equation}
    \mathcal{L}_{\text{ft}} = \mathcal{L}_{\text{task}} + \lambda \mathcal{L}_{\text{balance}},
\end{equation}
where $\mathcal{L}_{\text{task}}$ denotes the task-specific loss (e.g., $\mathcal{L}_{\text{seg}}$ for segmentation, $\mathcal{L}_{\text{det}}$ for detection).

\section{Experiments}
\label{sec:experiments}
\subsection{Implementation Details}
We conduct self-supervised learning on six mainstream datasets, including ScanNet \cite{dai2017scannet}, S3DIS \cite{s3dis}, Structured3D \cite{zheng2020structured3d}, 3D-Front \cite{3dfront}, ARKitScenes \cite{baruch2021arkitscenes} and HM3D \cite{hm3d}, comprising a total of 47,273 training samples. 
The network architecture follows the same design as Sonata, featuring 5 stages with block counts of 3, 3, 3, 12, and 3 per stage.
Pretraining configurations are as follows: batch size is set to 64, learning rate is 0.0004, and training spans 50 epochs. Following Sonata \cite{sonata}, patch masking employs a cosine scheduler. Furthermore, density variations and color dropout are applied with different sampling ratios across patches. More experimental details are provided in supplementary materials. 
For the DoReMi architecture, the maximum expert number $K$ is set to 8. $\lambda$ is set to 0.001. The encoder architecture closely follows the pretrained network, with Re and Do branches added only to the final block of each stage. Domain embeddings are randomly initialized. All experiments are conducted on 8 NVIDIA A100 GPUs, with AdamW optimizer employed throughout.

\begin{table}[t]
\small
\centering

\caption{\textbf{Outdoor semantic segmentation.} $^\star$ indicates results reproduced from the official repository.}
\begin{tabular}{l|ccc ccc}
\toprule
\multirow{2}{*}{Method} & \multicolumn{3}{c}{Nuscenes Val} & \multicolumn{3}{c}{Waymo Val} \\
\cline{2-4} \cline{5-7}
 & mIoU & mAcc & allAcc & mIoU & mAcc & allAcc \\
\hline
PTv3 \cite{ptv3}& 80.4 & 87.2 & 94.7 & 71.3 & 80.5 & 94.7 \\
Sonata$^\star$ \cite{sonata} & 81.2 & \textbf{87.7} & \textbf{94.8} & 72.1 & 82.6 & \textbf{94.8} \\
\rowcolor[HTML]{caeef0}DoReMi (Ours) & \textbf{81.7} & 87.4 & \textbf{94.8} & \textbf{72.7} & \textbf{82.7} & \textbf{94.8} \\
\bottomrule
\end{tabular}
\label{tab:outdoor_result}
\end{table}
\subsection{Experimental Results}
\noindent \textbf{Indoor Semantic Segmentation.}
Following PPT \cite{ppt}, we conduct joint training on three datasets: ScanNet (20 classes), S3DIS (13 classes), and Structured3D (25 classes). To achieve cross-dataset semantic alignment of categories, we incorporate a CLIP-based classification head. We directly evaluated performance on ScanNet and S3DIS. Furthermore, we fine-tune our model on the more challenging ScanNet200 benchmark, achieving SOTA performance across all three core metrics (mIoU, mAcc, and allAcc), as detailed in Table \ref{tab:results}. Specifically, our method attains mIoU scores of 80.1\%, 37.2\%, and 77.2\% respectively, representing improvements of 0.5\%, 0.4\%, and 1.2\% over Sonata. 
Figure \ref{fig:visualizaiton} presents a qualitative comparison with other methods. In the first scenario, given an incomplete chair, other approaches misidentify it as a sofa, whereas our method recognizes it as a chair. This demonstrates that the Re obtained through SSL significantly enhances the model's capability to extract discriminative features from incomplete objects. In the second scenario, while other methods erroneously classify a table as a desk, our approach leverages DSR to integrate surrounding spatial context information. This enables precise routing of inputs to specialized expert networks through spatial feature analysis, thereby achieving superior recognition performance by fully exploiting contextual dependencies and domain-specific knowledge. 
These results demonstrate DoReMi's superior cross-domain adaptability and generalization capability in complex indoor scenarios.





\noindent \textbf{Outdoor Semantic Segmentation. }
Following indoor experimental setup, we conduct joint training and direct evaluation on nuScenes \cite{caesar2020nuscenes} and Waymo \cite{waymo} datasets. As presented in Table \ref{tab:outdoor_result}, our model achieves mIoU scores of 81.7\% and 72.7\% on nuScenes and Waymo datasets respectively, representing 0.5\% and 0.6\% improvements over Sonata. These results validate DoReMi's superior generalization capability in outdoor scenarios.

\noindent \textbf{Extension to Unseen Scenes. }
Beyond standard benchmarks, we evaluate our jointly trained model on the unseen Matterport3D dataset \cite{chang2017matterport3d}, which comprises 90 large-scale indoor scenes. Using domain embeddings averaged from ScanNet, S3DIS, and Structured3D, we assess both fine-tuned and zero-shot generalization. Under fine-tuning (Table \ref{tab:matterport3d}), our model achieves 57.1\% mIoU, 68.2\% mAcc, and 83.3\% allAcc, outperforming Sonata by 1.0\%, 2.3\%, and 1.0\%, respectively. In the zero-shot setting—using the same model without further training—DoReMi also significantly surpasses Sonata and Point-MoE across all metrics (Table \ref{tab:dg_comparison}). This strong cross-domain generalization demonstrates that DSR generalizes through local spatial topology rather than memorizing domain embeddings. Even with averaged embeddings, the model effectively activates appropriate experts, indicating that the routing mechanism captures structural priors transferable to novel scenes.


\begin{table}[t]
    \small
    \centering
    \begin{minipage}[t]{0.50\linewidth}
        \centering
        \caption{\textbf{Extension to unseen scenes.} Segmentation results on Matterport3D val.}
        \label{tab:matterport3d}
            \begin{tabular}{lccc}
                \toprule
                Method & mIoU & mAcc & allAcc \\
                \midrule
                PTv3 \cite{ptv3}& 54.5 & 65.0 & 81.5 \\
                Sonata \cite{sonata}& 56.1 & 65.9 & 82.3 \\
                \rowcolor[HTML]{caeef0}DoReMi (Ours) & \textbf{57.1} & \textbf{68.2} & \textbf{83.3} \\
                \bottomrule
            \end{tabular}
    \end{minipage}
    \hfill 
    \begin{minipage}[t]{0.49\linewidth}
        \centering
        \caption{Zero-shot domain generalization comparison on Matterport3D val.}
        \label{tab:dg_comparison}
        \renewcommand{\arraystretch}{1.025}
            \begin{tabular}{lccc}
                \toprule
                Method & mIoU & mAcc & allAcc \\
                \midrule
                Sonata & 48.1 & 61.0 & 77.6 \\
                Point-MoE~\cite{chen2025point} & 41.8 & - & - \\
                \rowcolor[HTML]{caeef0} {DoReMi (Ours)} & \textbf{49.4} & \textbf{61.9} & \textbf{78.6} \\
                \bottomrule
            \end{tabular}
    \end{minipage}
\end{table}

\begin{table}[t]
\small
\centering
\caption{\textbf{Multimodal object detection} results on the ARKitScenes validation set.}

\begin{tabular}{l|cc}
\toprule
Method & F1@0.25 & F1@0.5 \\
\hline
SpatialLM \cite{SpatialLM} + Sonata \cite{sonata}& 58.9 & 49.5 \\
\rowcolor[HTML]{caeef0} SpatialLM \cite{SpatialLM} + DoReMi (Ours) & \textbf{60.8} & \textbf{51.9} \\
\bottomrule
\end{tabular}
\label{tab:detect_comparison}
\end{table}

\noindent \textbf{Multimodal Object Detection. }
Beyond segmentation, our method also extends effectively to detection tasks. In this experiment, we implement the SpatialLM \cite{SpatialLM} for object detection on the ARKitScenes dataset \cite{baruch2021arkitscenes}. 
Evaluation employs F1-score metrics under two IoU thresholds (0.25 and 0.5). We establish the baseline using officially fine-tuned SpatialLM model weights with Sonata-based point cloud encoder, which serves as the initialization for DoReMi's subsequent fine-tuning. The optimization employs a learning rate of 0.00005 over 10 epochs. As summarized in Table \ref{tab:detect_comparison}, our method achieves 60.8\% F1@0.25 and 51.9\% F1@0.5, outperforming the baseline by 1.9\% and 2.4\%. These results validate the cross-task effectiveness of our framework and demonstrate its potential for extension to diverse vision tasks.
\subsection{Ablation Studies}






\begin{table}[t]
\small
    \centering
    \begin{minipage}[t]{0.64\linewidth} 
        \centering
        \caption{\textbf{Ablation study on each component.}}
        \label{tab:ablation}
        
        \resizebox{\linewidth}{!}{ 
            \begin{tabular}{ccc|cccccc}
                \toprule
                \multirow{2}{*}{Re} & \multirow{2}{*}{DSR} & \multirow{2}{*}{EDA} & \multicolumn{3}{c}{ScanNet Val} & \multicolumn{3}{c}{S3DIS Val} \\
                \cline{4-6} \cline{7-9}
                 & & & mIoU & mAcc & allAcc & mIoU & mAcc & allAcc \\ 
                \hline 
                × & × & × & 77.5 & 85.4 & 92.1 & 73.5 & 78.9 & 92.0 \\
                $\checkmark$ & × & × & 78.8 & 85.9 & 92.5 & 75.7 & 81.4 & 92.5 \\
                $\checkmark$ & $\checkmark$ & × & 79.5 & 87.0 & 92.9 & 76.4 & 82.1 & 93.0 \\
                \rowcolor[HTML]{caeef0} $\checkmark$ & $\checkmark$ & $\checkmark$ & 80.1 & 87.3 & 93.1 & 77.2 & 82.5 & 93.1 \\
                \bottomrule
            \end{tabular}
        }
    \end{minipage}
    \hfill 
    \begin{minipage}[t]{0.34\linewidth} 
        \centering
        \caption{\textbf{Comparison of DoReMi layer placement positions.}}
        \label{tab:N_performance}
        
        \resizebox{\linewidth}{!}{ 
            \begin{tabular}{ccccc}
                \toprule
                 & $N_{MoE}$ & {mIoU} & {mAcc} & {allAcc} \\
                \midrule 
                \#1 & 3 & 79.4 & 86.8 & 92.7 \\
                \rowcolor[HTML]{caeef0}\#2 & 5 & 80.1 & 87.3 & 93.1 \\
                \#3 & 10 & 79.9 & 87.2 & 92.9 \\
                \bottomrule
            \end{tabular}
        }
    \end{minipage}
    
\end{table}
\noindent \textbf{Component Ablation. }
As shown in Table \ref{tab:ablation}, we conduct ablation studies on ScanNet and S3DIS to validate the effectiveness of each core component in DoReMi framework. 
First, when only Re branch is enabled, the model achieves improvements of 1.3\% and 2.2\% in mIoU compared to the baseline method PPT, demonstrating that Re branch enhances the model's ability to capture shared semantic patterns in multi-domain data through generic pretrained knowledge. 
Further incorporating DSR, with two experts selected, yields an additional 0.7\% gain on both datasets. This improvement verifies that DSR can dynamically select the most relevant expert subsets based on the domain characteristics of input samples, thereby enhancing the model's adaptability and generalization capability in multi-domain scenarios. 
Finally, when EDA is integrated, the model achieves optimal performance with mIoU scores of 80.1\% and 77.2\%. EDA further improves expert utilization efficiency and training stability by optimizing expert load balancing and activation counts. 
These results validate the synergy within DoReMi: the Re branch establishes robust, domain-invariant representations, while DSR and EDA enable adaptive routing via topological awareness and uncertainty optimization. This facilitates dynamic activation of optimal combination of experts to bridge complex distribution shifts, creating a powerful and scalable framework for multi-domain point cloud understanding. 





\begin{table}[t]
\small
    \centering
    \begin{minipage}[t]{0.64\linewidth}
        \centering
        \caption{\textbf{Expert Number.} Evaluation of performance variations under different expert numbers.}
        \label{tab:moe_num}
        
        {
            \begin{tabular}{cccc}
                \toprule
                Num & {mIoU} & {mAcc} & {allAcc} \\
                \midrule 
                4  & 79.6 & 87.2 & 92.8 \\
                6  & 79.7 & 86.9 & 93.0 \\
                \rowcolor[HTML]{caeef0}8  & 80.1 & 87.3 & 93.1 \\
                10 & 79.5 & 87.1 & 92.9 \\
                12 & 79.4 & 86.6 & 92.7 \\
                \bottomrule
            \end{tabular}
        }
    \end{minipage}
    \hfill 
    \begin{minipage}[t]{0.34\linewidth}
        \centering
        \caption{\textbf{Evaluation of expert load balancing performance.} We use the normalized standard deviation as the metric.}
        \label{tab:load_balance}
        
        {
            \begin{tabular}{ccc}
                \toprule
                Method & {$\alpha \downarrow$} & mIoU\\
                \midrule 
                Vanilla MoE &  0.941 &  77.3 \\
                \rowcolor[HTML]{caeef0}DoReMi (Ours) &  0.894 &80.1 \\
                \bottomrule
            \end{tabular}
        }
    \end{minipage}
    
\end{table}





\noindent \textbf{MoE-specific Analysis. }
We analyze the design of experts in DoReMi Layer and its load balancing. Table \ref{tab:N_performance} shows the impact of inserting DoReMi layers at different encoder stages. Specifically, \#1 inserts them at the final Block of stage 1, 3, and 5, totaling 3 insertion points; \#2 inserts them at the final Block of every stage, totaling 5 insertion points; \#3 inserts them at both the first and final Blocks of each stage, totaling 10 insertion points. Experimental results indicate that \#2 achieves the best performance, reaching a performance metric of 80.1\%. Insufficient insertion points (e.g., \#1) limit the model's ability to adequately capture hierarchical feature representations, while excessive insertions (e.g., \#3) may introduce redundancy. 
Table \ref{tab:moe_num} analyzes the effect of the number of experts in the Do branch. Experiments show that using 8 experts yields the best performance. Fewer experts (e.g., 4) limit model expressiveness, while more experts increase complexity, training instability, and overfitting risk. 
Table \ref{tab:load_balance} evaluates load balancing using the normalized standard deviation of expert activation counts, denoted as:
\begin{equation}
    \alpha = \frac{\sqrt{\frac{1}{K}\sum_{j=1}^{K}(c_j - \bar{c})^2}}{\bar{c}},
\end{equation}
where $\bar{c} = \frac{1}{K}\sum_{j=1}^{K}c_j $. In ``Vanilla MoE'', 8 experts are configured with 2 experts fixed for activation, while our method employs EDA. Experimental results show that our method achieves a lower normalized standard deviation compared to ``Vanilla MoE'', indicating that our mechanism distributes expert load more evenly, achieving superior load balancing performance and effectively improving  model stability.



\begin{figure}[t]
    \centering
    \begin{minipage}[t]{0.48\linewidth}
        \vspace{0pt} 
        \centering
        \makeatletter\def\@captype{table}\makeatother 
        
        \caption{\textbf{Efficiency comparison.} ${'}$ and ${''}$ represent Sonata variant. ``Act. Params'' indicates the number of activated parameters. FPS is measured on one A100 GPU.}
        \label{tab:efficiency}
        
        \resizebox{1\linewidth}{!}{
        \begin{tabular}{c|l|ccc}
        \toprule
        & Method & Act. Params & FPS & mIoU\\
        \hline
        \#1 &PTv3  \cite{ptv3} & 46.2M  & 11.0 & 77.6 \\
        \#2 &PPT  \cite{ppt} & 46.3M  & 7.2 & 78.6\\
        \#3 &Sonata \cite{sonata}& 124.8M & 5.9 & 79.4 \\
        \#4 &Sonata${'}$ & 147.6M & 5.3 & 79.4 \\
        \#5 &Sonata${''}$ & 148.6M & 4.9 & 79.6 \\
        \rowcolor[HTML]{caeef0} \#6 & DoReMi (Ours)   & 147.5M & 4.9  & 80.1\\
        \bottomrule
        \end{tabular}
        }
    \end{minipage}
    \hfill
    \begin{minipage}[t]{0.50\linewidth}
        \vspace{0pt} 
        \centering
        \includegraphics[width=1\linewidth]{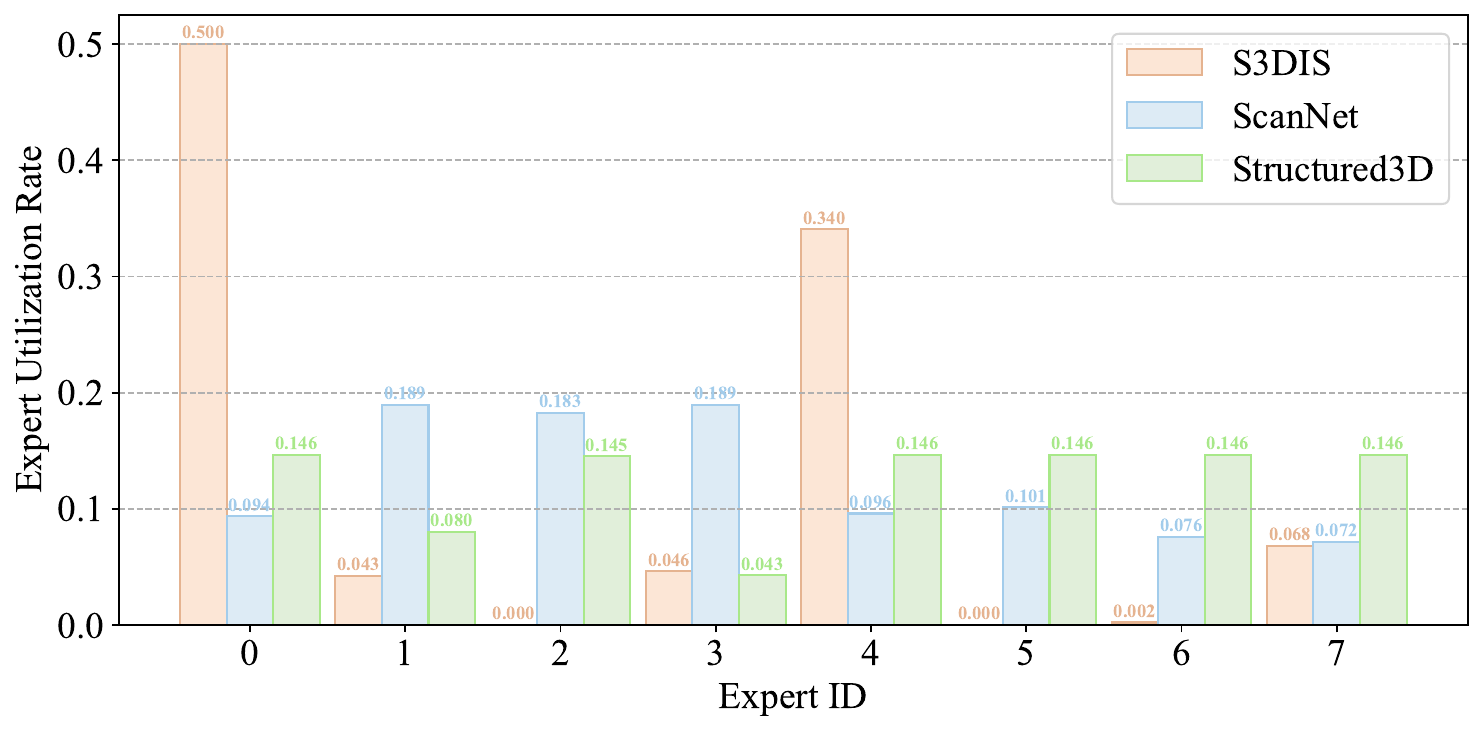}
        \caption{\textbf{Expert utilization rates across different datasets.} Taking Encoder Stage 2 as an example, y-axis values represent dataset-specific normalized utilization.}
        \label{fig:expert_usage}
    \end{minipage}
\end{figure}

\noindent \textbf{Efficiency Comparison. }
We further analyze the efficiency–accuracy trade-off of DoReMi and previous methods.
As shown in Table~\ref{tab:efficiency}, Sonata~\cite{sonata} achieves higher performance than PPT \cite{ppt} but at the cost of a large activated parameter increase.
In contrast, our DoReMi further improves mIoU over Sonata (+0.7\%) with only a moderate parameter growth.
To verify that the performance gains are not merely due to model scaling, we implement two larger variants of Sonata. The first sets the stage depths to [2, 2, 6, 2] and expands the channel dimensions to [64, 128, 256, 512], as shown in \#4, while the second increases the depth of all four decoder stages to 5, as shown in \#5. Both variants have a comparable number of parameters to DoReMi and follow the official implementation and training settings. Despite having similar model capacity, these variants still underperform compared to DoReMi, demonstrating that the observed improvements primarily stem from our domain–representation mixture design rather than parameter scaling.
Overall, DoReMi delivers a superior accuracy–efficiency balance, converting limited additional computation into consistent improvements.

\noindent \textbf{Expert Activation Analysis. }
Figure \ref{fig:expert_usage} shows the activation frequency of Do branch experts in the second DoReMi layer across domains, with additional results in the supplementary material. In S3DIS, Experts 0 and 4 are activated more frequently, suggesting they specialize in the dataset’s spatial and semantic patterns. In Structured3D, activations are more balanced due to diverse scene types, allowing the model to leverage multiple experts for robust generalization. 
This demonstrates that our routing mechanism enables the model to adaptively activate different combinations of experts based on the distinct topological structures across various datasets.

\begin{figure}[t]
    \centering
    \begin{minipage}[t]{0.48\linewidth}
        \vspace{0pt} 
        \centering
        \includegraphics[width=1\linewidth]{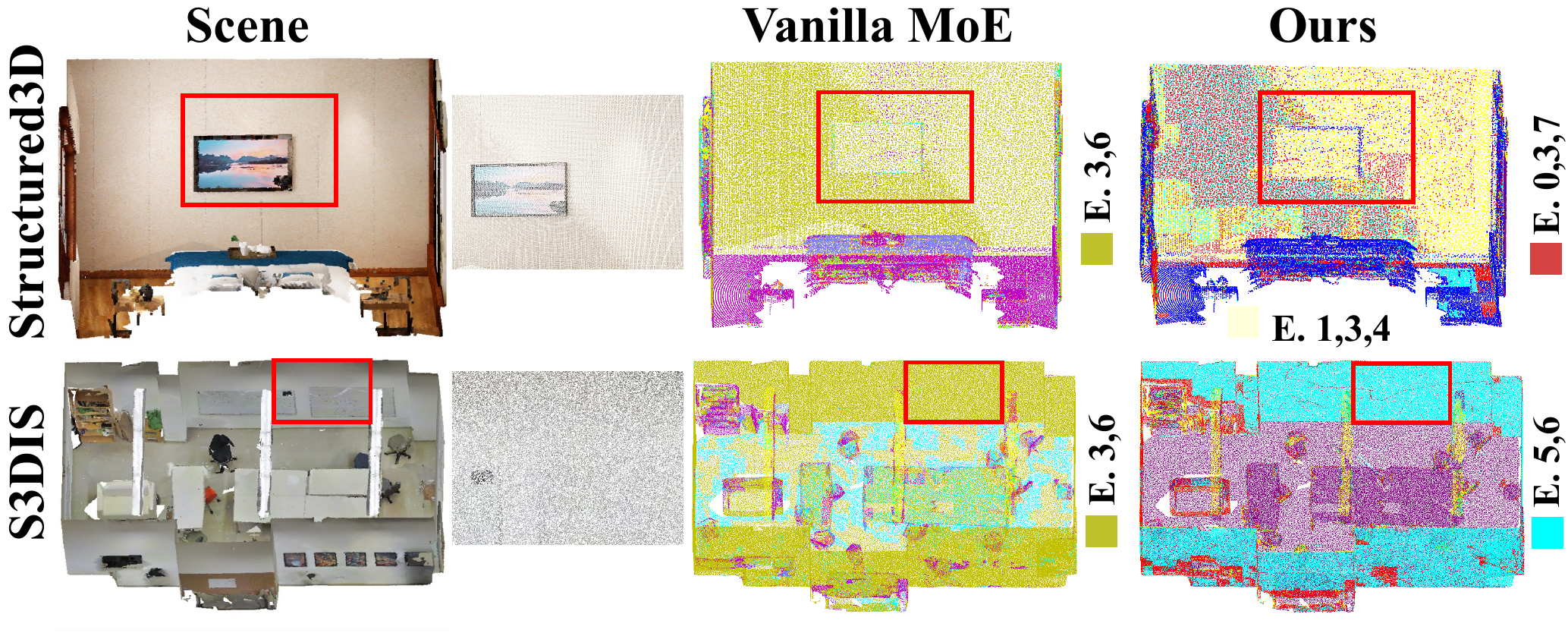}
        \caption{Semantic-dominant vs. Topology-aware routing across different datasets.}
        \label{fig:fig6}
    \end{minipage}
    \hfill
    \begin{minipage}[t]{0.50\linewidth}
        \vspace{0pt} 
        \centering
        \makeatletter\def\@captype{table}\makeatother 
        
        \caption{\textbf{Comparison of different pretraining strategies.}}
        \label{tab:pretrain_compare}
        
        \resizebox{1\linewidth}{!}{
        \begin{tabular}{c|l| >{\centering\arraybackslash}p{1.1cm}|>{\centering\arraybackslash}p{1.1cm} |ccc}
        \toprule
            & \multicolumn{1}{c|}{\multirow{2}{*}{Method}} & \multicolumn{2}{c|}{Visual Modality} & \multicolumn{1}{c}{\multirow{2}{*}{mIoU}} & \multicolumn{1}{c}{\multirow{2}{*}{mAcc}} & \multicolumn{1}{c}{\multirow{2}{*}{allAcc}} \\ \cline{3-4}
            & \multicolumn{1}{c|}{}                        & \multicolumn{1}{c}{2D}                & \multicolumn{1}{c|}{3D}               & \multicolumn{1}{c}{}                      & \multicolumn{1}{c}{}                      & \multicolumn{1}{c}{}                        \\ \hline
        \#1 & Sonata     \cite{sonata}         &    ×    & \checkmark   & 79.4     & 86.1    & 92.5        \\
        \#2 & Sonata+Ours  &   ×  & \checkmark     & 79.7   &  87.0   &  92.8   \\
        \#3 & DoReMi (Ours)     &  ×   & \checkmark & 80.1   & 87.3  & 93.1        \\
        \#4 & Concerto  \cite{zhang2025concerto}    & \checkmark  & \checkmark & 80.7  & 87.4     & 93.1   \\
        \#5 & Concerto+Ours  & \checkmark  & \checkmark  & 81.2  & 89.1    & 93.2      \\ 
        \bottomrule      
        \end{tabular}
        }
    \end{minipage}
\end{figure}

As discussed in Section~\ref{sec:intro} and demonstrated in Fig.~\ref{fig:teaser}, vanilla 3D MoE (following Point-MoE~\cite{chen2025point}) suffer from semantic-dominant routing. Under intra-scene variations, even when identical objects (e.g., a "bed" and "floor") present local topological fluctuations due to occlusion or scanning patterns, Vanilla MoE rigidly assigns the exact same experts. As shown in Fig.~\ref{fig:fig6}, this limitation of Vanilla MoE extends to cross-dataset scenarios: for the same semantic category like a "wall", it activates identical experts, ignoring the distinct point cloud topological characteristics caused by different acquisition methods across datasets. In contrast, our approach assigns distinct expert combinations to adaptively handle different topological structures.

\noindent \textbf{Effectiveness of Pretraining Strategies. }
To validate the adaptability and effectiveness of the DoReMi framework under different pretrained representations, we present in Table \ref{tab:pretrain_compare} the experimental results obtained by initializing the model with various pretraining strategies.
The comparison between \#1 and \#2 demonstrates that, under identical pretrained weights, incorporating the structural design of DoReMi significantly improves model performance, indicating its ability to effectively integrate and enhance existing representations.
The contrast between \#2 and \#3 shows that, when pretrained solely on point cloud modality, our multi-attribute self-supervised learning achieves better results, suggesting that richer pretraining signals enable DoReMi to better preserve cross-domain geometric and structural priors, thereby improving generalization.
Recently, Concerto \cite{zhang2025concerto} introduced 2D–3D joint self-supervised learning, which leverages additional image modalities to obtain stronger representations. Combining its pretrained weights with the DoReMi architecture further pushes the state-of-the-art performance, demonstrating that DoReMi’s structural design is compatible with various pretraining paradigms and can effectively enhance performance on top of strong feature representations.




\section{Conclusion}
This paper proposes DoReMi, a novel Domain-Representation Mixture-of-Experts framework for 3D scene understanding. To address the issue of ``semantically consistent yet topologically heterogeneous" data caused by sensor heterogeneity in multi-source 3D datasets, DoReMi employs a frozen Unified Representation branch (Re). This branch leverages multi-attribute self-supervised learning to establish robust cross-domain structural priors. Meanwhile, the Domain-aware branch (Do) introduces Domain-Spatial guided Routing (DSR) and an Entropy-controlled Dynamic Allocation (EDA) mechanism. These components overcome the limitations of traditional MoE models that rely on semantic-guided routing, enabling dynamic perception of fine-grained physical topological variations and adaptive expert allocation.
Extensive experiments on multiple indoor and outdoor 3D understanding benchmarks demonstrate that DoReMi effectively manages drastic geometric fluctuations in density and completeness, showcasing superior cross-domain adaptability and generalization performance.

%
%
\bibliographystyle{splncs04}
\bibliography{main}
\end{document}